\documentclass{article}
\usepackage[numbers]{natbib}
\usepackage[final]{neurips_2023} 

\usepackage[utf8]{inputenc} 
\usepackage[T1]{fontenc} 
\usepackage{url} 
\usepackage{booktabs} 
\usepackage{amsfonts} 
\usepackage{nicefrac} 
\usepackage{microtype} 
\usepackage{xcolor} 

\usepackage{graphicx}

\begin{document}

\author{%
Amritpal Singh\thanks{Correspondence Email: asingh880@gatech.edu \\
Accepted in Medical Imaging meets NeurIPS Workshop, 2023.} \\
  Georgia Institute of Technology, USA\\
  \texttt{asingh880@gatech.edu} \\
  \And
  Mustafa Burak Gurbuz \\
   Georgia Institute of Technology, USA\\
  \texttt{mgurbuz6@gatech.edu}  \\
    \And
  Shiva Souhith Gantha \\
   Georgia Institute of Technology, USA\\
  \texttt{sgantha3@gatech.edu}  \\
    \And
  Prahlad Jasti \\
   Georgia Institute of Technology, USA\\
   \texttt{pjasti3@gatech.edu}  \\
}

\title{Class-Incremental Continual Learning for General
Purpose Healthcare Models}

\maketitle 

\begin{abstract}
Healthcare clinics regularly encounter dynamic data that changes due to variations in patient populations, treatment policies, medical devices, and emerging disease patterns. Deep learning models can suffer from catastrophic forgetting when fine-tuned in such scenarios, causing poor performance on previously learned tasks. Continual learning allows learning on new tasks without performance drop on previous tasks. In this work, we investigate the performance of continual learning models on four different medical imaging scenarios involving ten classification datasets from diverse modalities, clinical specialties, and hospitals. We implement various continual learning approaches and evaluate their performance in these scenarios. Our results demonstrate that a single model can sequentially learn new tasks from different specialties and achieve comparable performance to naive methods. These findings indicate the feasibility of recycling or sharing models across the same or different medical specialties, offering another step towards the development of general-purpose medical imaging AI that can be shared across institutions.
\end{abstract} 

\section{Introduction} 
Deep Neural Networks (DNNs) have recently exhibited remarkable achievements in various tasks, surpassing human expertise in some cases \citep{visualUnderstanting, deepLearning, alphaGo}. However, their dependence on fixed, balanced datasets within stable environments presents a significant constraint. The ever-changing nature of the real world requires networks capable of sequential learning over time and adapting to shifting data distributions. This shortcoming is especially pronounced in healthcare and medical imaging. The emergence of new diseases, changes in patient population, treatment policies, disease distribution, imaging hardware, or image acquisition techniques can significantly impact the model's performance. Fine-tuning exclusively on new data, adapts models to the latest targets, resulting in a rapid loss of previously acquired knowledge. Techniques such as Joint Training (JT) are used to overcome this, where the model is trained on both old and new data. However, healthcare data can't always be shared due to safety concerns and regulation differences across geographies. On the contrary, the NAIVE approach trains an independent model for each task, increasing computational resources required for training, deployment, and missing performance gain due to shared representation.
Continually learning is an active area of research, allowing efficient training and adaptation of algorithms to new data without losing prior knowledge. This can improve the model updating, sharing, and resource optimization for healthcare institutions. Cross-sharing models between multiple hospitals can benefit both institutions. Finally, healthcare professionals will be able to screen and detect patients more effectively by quickly identifying and analyzing new biomarkers as they emerge with disease or population shift. 
 
In this work, we show the feasibility of training continually learning models for medical imaging. Our contributions are as follows:

\begin{enumerate} 
 \item We explore the potential of continual learning for sharing medical imaging AI algorithms across changes in hospitals/geographies, medical specialties, and imaging modalities. 
 \item We create 4 continual learning scenarios to assess cross-sharing (inter-hospital scenario and one inter-specialty) and intra-specialty model recycling (pathology, radiology) use cases. 
 \item We show that continual learning methods can gain performance on par with naive and joint learning approaches while remembering previous tasks. 
\end{enumerate} 

\section{Methods} 
We implement 6 variants of continual learning methods, namely memory aware synapses (MAS), replay using memory indexing (REMIND), MAS with replay (MAS+r), Neuro-inspired stability-plasticity adaptation (NISPA), dark experience replay (DER), and DER++; with prior two not requiring data access to previous tasks. The final four require data access, generally solved by local storage in replay buffers of fixed sizes.
MAS \citep{MAS} and its replay variant MAS+r are regularization methods that calculate the importance of the model parameters in an online fashion. REMIND \citep{remind} stores compressed low-level feature representations instead of actual input data, making it well-suited when past data is not feasible. NISPA \citep{nispa} uses a rewiring mechanism inspired by the structural plasticity of biological neurons and driven by local activations of units similar to Hebbian learning in the brain. DER \citep{DER} and DER++ are replay methods that selectively choose examples with high uncertainty to replay. 

We train naïve learner, and joint learner for baseline comparison. The naive learner corresponds to training an independent model for each data, and the joint learner is trained on all data together. We use a 5-layered convolutional neural network (CNN) followed by a linear layer classifier as a backbone model. 

No task labels are provided during testing. To succeed, the model needs to learn inter-task differences to predict task ID on testing and learn intra-task differences to predict the correct class. We measure task accuracy percentage after every episode, average accuracy on seen classes after completing an episode, and backward transfer. 

\begin{figure} 
 \centering 
 \includegraphics[width=\linewidth]{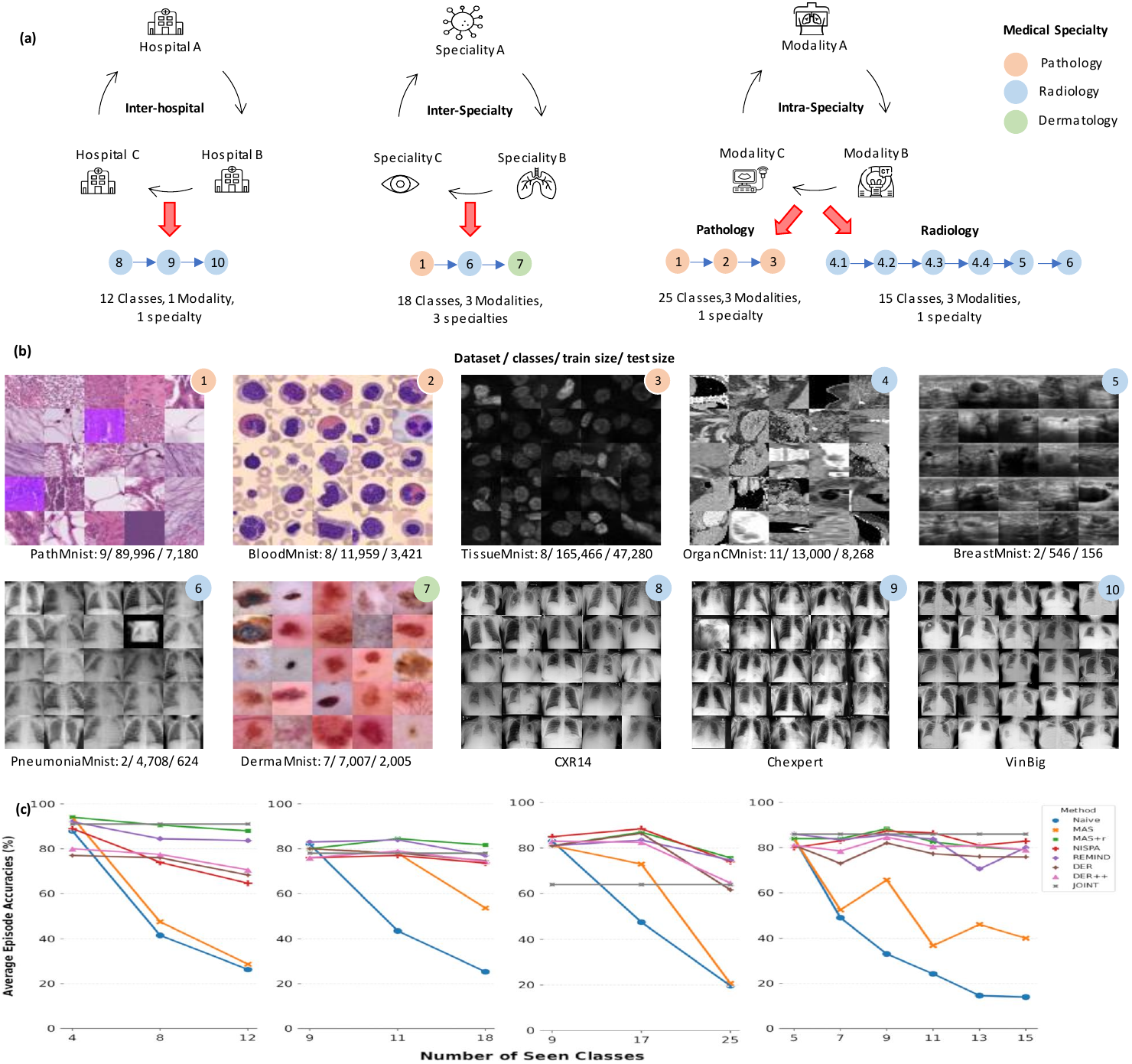} 
 \caption{Different learning scenarios and dataset. (a) shows scenarios and sequences of learning tasks used to train the algorithms. (b) shows snapshots of images from different datasets, color-coded (as per their specialty) sequence index on the right top of images. Each dataset has two or more classification labels. Note the visual similarities inside each dataset and the diversity among the different tasks in the scenario. Dataset 4 is split into 4 sub-tasks with unique classes, named 4.1, 4.2, 4.3 and 4.4} 
 \label{fig:results_acc} 
\end{figure}

\subsection{Datasets and scenarios}

Figure \ref{fig:results_acc} provides a snapshot of scenarios and datasets used. We devise four continual learning scenarios, divided into 3 major categories, to simulate learning new tasks inside the same specialty, different specialties, and hospitals. 

\textbf{Inter-hospital scenario} simulates model sharing across hospitals from different geographies. We used x-rays for pleural effusion, cardiomegaly, atelectasis, and consolidation from the Chexpert dataset, CXR-14 dataset, and VinBig dataset; in total representing 3 hospitals and 2 countries. \textbf{Inter-Specialty scenarios} simulate model sharing between different medical specialties inside the same hospital and can help specialties with fewer data to benefit from those with more. We combine three specialties; pathology\citep{pathmnist}, radiology\citep{pneumoniamnist}, and dermatology\citep{dermamnist}. \textbf{Intra-Specialty scenarios} simulate model rotation within a specialty, mimicking learning new disease finding that appears later on. The pathology scenario contains three subtasks: histology of colorectal cancer \citep{pneumoniamnist}, blood cells \citep{bloodmnist}, and kidney cortex cells \citep{tissuemnist}. The radiological scenario contains three subtasks: computed tomography (CT) \citep{organcmnist}), ultrasound \citep{breastmnist}, and chest X-ray \citep{pneumoniamnist} images.

\section{Results and Discussion} 
In this section, we discuss the performance of continual learning methods with comparison at the scenario and algorithm level. Figure \ref{fig:results_acc} shows the average accuracy of methods on test data, with every point representing average accuracy on current and all previous tasks. As expected accuracy for NAIVE method takes a sharp dip. Continual learning methods perform on par or better, with NAIVE on the current task, with little to no drop in their performance on the previous task. Across all continual learning methods, replay methods perform better than regularization methods across all scenarios. MAS with replay and DER++ get high accuracies compared to other methods. MAS+r is consistently the best performer in all scenarios, achieving an average accuracy of 88, 82, 75, 79, on inter-hospital, inter-specialty, pathology, and radiology respectively at the end of each scenario. MAS with replay had the highest backward transfer with a value of -2, -5, +3, and -5 on the respective scenarios. Among non-replay methods, MAS has a huge performance drop. REMIND achieves an average accuracy of 83, 77, 75, 80 on inter-hospital, inter-specialty, pathology, and radiology respectively.
It's important to note that replay methods have access to a subset of previous data stored for future retraining. On the other hand, REMIND stores compressed representations, instead of the actual data, allowing memory efficiency and data privacy. Another critical point to note is REMIND's dependence on its feature extractor. Since the initial feature extractor is frozen after its initialization, the model is less flexible for learning tasks unrelated to initial studies. Using pre-trained weights from bigger datasets can provide a boost in model performance. 

\textbf{Limitations and Future work}
We don't explore the effect of the sequence of tasks on learning, as this can impact quality of features learned. We used 32x32x3 image size on a small CNN architecture. Using higher resolution medical imaging and larger pre-trained models can help boost performance further. This can be an important future direction.

\section{Conclusion} 
In this work, we explore the performance of continuous learning methods in varying specialties, modalities, and geographies. We show that continual learning algorithms can learn new tasks while maintaining performance on previous tasks, even while changing modalities specialties, and hospitals. This shows potential in developing general-purpose medical imaging AI that can be shared across institutions, with the ability to adapt to new tasks. 

\section{Potential Negative Societal Impacts}
Continual learning models may inherit biases present in the data on which they are trained. If the training data is not representative, these biases can lead to disparities in medical diagnoses and treatment recommendations. While these risks are inherent in deep learning models, automatic unsupervised training of continual learning can exacerbate these biases when deployed, often going unnoticed.
Furthermore, due to concerns related to quality control and disparities in deployment regions, continual learning models may inadvertently generate incorrect or misleading medical images or interpretations, which can have detrimental consequences for patients if not adequately monitored and controlled. AI systems are highly sensitive, potentially leading to overdiagnosis and overtreatment if not properly calibrated, thereby resulting in unnecessary medical interventions and increased healthcare costs.
Lastly, as with any technological advancement, AI systems are susceptible to hacking and cybersecurity threats. Breaches of medical AI systems could result in unauthorized access to sensitive patient data or manipulation of diagnostic results, posing significant privacy and security risks.
 
{\small
\bibliographystyle{plain}
\bibliography{egbib2}

\begin{thebibliography}{10}

\bibitem{bloodmnist}
Andrea Acevedo, Anna Merino, Santiago Alferez, Angel Molina, Laura Boldu, and
  José Rodellar.
\newblock A dataset of microscopic peripheral blood cell images for development
  of automatic recognition systems.
\newblock {\em Data in brief}, 30:105474, jun 2020.

\bibitem{breastmnist}
Walid Al-Dhabyani, Mohammed Gomaa, Hussien Khaled, and Aly Fahmy.
\newblock Dataset of breast ultrasound images.
\newblock {\em Data in brief}, 28:104863, feb 2020.

\bibitem{MAS}
Rahaf Aljundi, Francesca Babiloni, Mohamed Elhoseiny, Marcus Rohrbach, and
  Tinne Tuytelaars.
\newblock Memory aware synapses: Learning what (not) to forget.
\newblock In {\em The European Conference on Computer Vision (ECCV)}, 2018.

\bibitem{DER}
Pietro Buzzega, Matteo Boschini, Angelo Porrello, Davide Abati, and Simone
  Calderara.
\newblock Dark experience for general continual learning: a strong, simple
  baseline.
\newblock In {\em Advances in Neural Information Processing Systems},
  volume~33, 2020.

\bibitem{dermamnist}
Noel C.~F. Codella, Veronica Rotemberg, Philipp Tschandl, M.~Emre Celebi,
  Stephen~W. Dusza, David~A. Gutman, Brian Helba, Aadi Kalloo, Konstantinos
  Liopyris, Michael~A. Marchetti, Harald Kittler, and Allan Halpern.
\newblock Skin lesion analysis toward melanoma detection 2018: {A} challenge
  hosted by the international skin imaging collaboration {(ISIC)}.
\newblock {\em CoRR}, abs/1902.03368, 2019.

\bibitem{visualUnderstanting}
Yanming Guo, Yu~Liu, Ard Oerlemans, Songyang Lao, Song Wu, and Michael~S. Lew.
\newblock Deep learning for visual understanding: A review.
\newblock {\em Neurocomputing}, 187, 2016.

\bibitem{nispa}
Mustafa~Burak Gurbuz and Constantine Dovrolis.
\newblock Nispa: Neuro-inspired stability-plasticity adaptation for continual
  learning in sparse networks.
\newblock {\em arXiv preprint arXiv:2206.09117}, 2022.

\bibitem{remind}
Tyler~L Hayes, Kushal Kafle, Robik Shrestha, Manoj Acharya, and Christopher
  Kanan.
\newblock Remind your neural network to prevent catastrophic forgetting.
\newblock In {\em Computer Vision--ECCV 2020: 16th European Conference,
  Glasgow, UK, August 23--28, 2020, Proceedings, Part VIII 16}, pages 466--483.
  Springer, 2020.

\bibitem{pathmnist}
Jakob~Nikolas Kather, Johannes Krisam, Pornpimol Charoentong, Tom Luedde,
  Esther Herpel, Cleo-Aron Weis, Timo Gaiser, Alexander Marx, Nektarios~A
  Valous, Dyke Ferber, Lina Jansen, Constantino~Carlos Reyes-Aldasoro, Inka
  Zörnig, Dirk Jäger, Hermann Brenner, Jenny Chang-Claude, Michael
  Hoffmeister, and Niels Halama.
\newblock Predicting survival from colorectal cancer histology slides using
  deep learning: A retrospective multicenter study.
\newblock {\em {PLoS} Medicine}, 16(1):e1002730, jan 2019.

\bibitem{pneumoniamnist}
Daniel~S Kermany, Michael Goldbaum, Wenjia Cai, Carolina C~S Valentim, Huiying
  Liang, Sally~L Baxter, Alex {McKeown}, Ge~Yang, Xiaokang Wu, Fangbing Yan,
  Justin Dong, Made~K Prasadha, Jacqueline Pei, Magdalene Y~L Ting, Jie Zhu,
  Christina Li, Sierra Hewett, Jason Dong, Ian Ziyar, Alexander Shi, Runze
  Zhang, Lianghong Zheng, Rui Hou, William Shi, Xin Fu, Yaou Duan, Viet A~N
  Huu, Cindy Wen, Edward~D Zhang, Charlotte~L Zhang, Oulan Li, Xiaobo Wang,
  Michael~A Singer, Xiaodong Sun, Jie Xu, Ali Tafreshi, M~Anthony Lewis, Huimin
  Xia, and Kang Zhang.
\newblock Identifying medical diagnoses and treatable diseases by image-based
  deep learning.
\newblock {\em Cell}, 172(5):1122--1131.e9, feb 2018.

\bibitem{deepLearning}
Yann LeCun, Y.~Bengio, and Geoffrey Hinton.
\newblock Deep learning.
\newblock {\em Nature}, 521, 2015.

\bibitem{tissuemnist}
Vebjorn Ljosa, Katherine~L Sokolnicki, and Anne~E Carpenter.
\newblock Annotated high-throughput microscopy image sets for validation.
\newblock {\em Nature Methods}, 9(7):637, jun 2012.

\bibitem{alphaGo}
David Silver, Aja Huang, Christopher Maddison, Arthur Guez, Laurent Sifre,
  George Driessche, Julian Schrittwieser, Ioannis Antonoglou, Veda
  Panneershelvam, Marc Lanctot, Sander Dieleman, Dominik Grewe, John Nham, Nal
  Kalchbrenner, Ilya Sutskever, Timothy Lillicrap, Madeleine Leach, Koray
  Kavukcuoglu, Thore Graepel, and Demis Hassabis.
\newblock Mastering the game of go with deep neural networks and tree search.
\newblock {\em Nature}, 529, 2016.

\bibitem{organcmnist}
Xuanang Xu, Fugen Zhou, Bo~Liu, Dongshan Fu, and Xiangzhi Bai.
\newblock Efficient multiple organ localization in {CT} image using {3D} region
  proposal network.
\newblock {\em {IEEE} transactions on medical imaging}, jan 2019.

\end{thebibliography}
}
\end{document}